\documentclass[conference]{IEEEtran}
\IEEEoverridecommandlockouts
\usepackage{cite}
\usepackage{amsmath,amssymb,amsfonts}
\usepackage{algorithmic}
\usepackage{graphicx}
\usepackage{textcomp}
\usepackage{xcolor}
\usepackage{subcaption}

\def\BibTeX{{\rm B\kern-.05em{\sc i\kern-.025em b}\kern-.08em
    T\kern-.1667em\lower.7ex\hbox{E}\kern-.125emX}}
\begin{document}

\title{Automated Playtesting of Matching Tile Games
}

\author{\IEEEauthorblockN{Luvneesh Mugrai}
\IEEEauthorblockA{\textit{Game Innovation Lab} \\
\textit{New York University}\\
New York City, US \\
lm3300@nyu.edu}
\and
\IEEEauthorblockN{Fernando Silva}
\IEEEauthorblockA{\textit{Independent Researcher} \\
Rio de Janeiro, Brazil \\
fms2005@gmail.com}
\and
\IEEEauthorblockN{Christoffer Holmg\r{a}rd}
\IEEEauthorblockA{\textit{modl.ai} \\
Copenhagen, Denmark \\
christoffer@holmgard.org}
\and
\IEEEauthorblockN{Julian Togelius}
\IEEEauthorblockA{\textit{Game Innovation Lab} \\
\textit{New York University}\\
New York City, US \\
julian@togelius.com}
}
\maketitle

\begin{abstract}
Matching tile games are an extremely popular game genre. Arguably the most popular iteration, Match-3 games, are simple to understand puzzle games, making them great benchmarks for research. In this paper, we propose developing different procedural personas for Match-3 games in order to approximate different human playstyles to create an automated playtesting system. The procedural personas are realized through evolving the utility function for the Monte Carlo Tree Search agent. We compare the performance and results of the evolution agents with the standard Vanilla Monte Carlo Tree Search implementation as well as to a random move-selection agent. We then observe the impacts on both the game's design and the game design process. Lastly, a user study is performed to compare the agents to human play traces.
\end{abstract}

\begin{IEEEkeywords}
Procedural Personas, Monte Carlo Tree Search, Genetic Evolution, Match-3
\end{IEEEkeywords}

\section{Introduction}
When playing a game with a strategic element, players have various approaches to go about trying to solve the level. Some of these approaches include maximizing the overall score after a certain number of moves, maximizing the number of possible moves, prioritizing making moves in specific regions of the game board, and prioritizing making a specific move type over other currently available move types (e.g. making a horizontal move over a vertical move). Specific player personas can then be further categorized into groups such as a long term planner and short term planner.

In this paper, we explore different methods of modeling player personas through evolving standard Vanilla Monte Carlo Tree Search (MCTS). We attempt to approximate different styles that human players would have when playing Match-3 games. Our objective of the experiments and paper is to develop four procedural personas, which model four different types of playstyles:
\begin{enumerate}
\item Trying to maximize score (Referred to as Agent MaxS)
\item Trying to minimize score (Referred to as Agent MinS)
\item Trying to maximize the available number of moves (Referred to as Agent MaxM)
\item Trying to minimize the available number of moves (Referred to as Agent MinM)
\end{enumerate}
Agents MaxS and MinS mimic the long term planner and the short term  player, respectively, while Agent MaxM mimics the persona of setting up the board for a multitude of possibilities. Agent MinM can be seen as the counterpart to this persona encompassed in Agent MaxM. 

Being able to model human players and playstyles opens the possibility of playtesting new levels, analyzing the approaches and how players play various levels for Match-3 games. Game designers would be able to gain further insights on various interaction patterns and study how various categories of players would respond within the Match-3 genre. The approach can also open up the ability to then observe and analyze the impacts of game design following the playstyle perspectives from the different agents.

\section{Background}
Matching tile games are a continuously popular game genre, dating back to games as early as Chain Shot and Tetris in 1985, and are currently now associated with the channel of casual, downloadable games. Development of such games follows a continuous process of sequential releases of games, with new levels being released over time. For the purposes of the experiments we will be focusing on games similar in nature to that of Bejeweled, developed by PopCap Games, and Candy Crush Saga, developed by King.

The approaches in this paper draw inspiration from Holmg\r{a}rd's work on \textit{Automated Playtesting with Procedural Personas through MCTS with Evolved Heuristics} \cite{Holmgard2018automated} and on \textit{Evolving Personas for Player Decision Modeling} \cite{Holmgard2014evolving}. Personas as a concept originally refers to hand-coded models, though procedural personas are often defined via evolution or reinforcement learning based on logs of play data ~\cite{Holmgard2018automated,tastan2011learning}. Previous work has shown the promising results of being able to use evolutionary methods in conjunction with MCTS to create personas for turn-based games \cite{Holmgard2018automated} \cite{Holmgard2014evolving}. The idea of procedural personas traces back to the term of \textit{play personas}, coined by Canossa and Drachen~\cite{Alessandro2008Defining}. It is used to better define personas in terms of how players chose to interact within the space of a game \cite{Alessandro2009patterns}. Procedural personas built on this idea but through computational, generative models. Generally, a procedural persona is defined in terms of utility functions and computational resources \cite{Holmgard2018automated} \cite{holmgard2014generative}. These personas could be implemented as agents to then re-create game-play interactions similar to those of different human player types. 

Monte Carlo methods are a class of algorithms that aim to solve a problem by sampling random values and approximating the mathematical property behind said problem. They are widely adopted in a wide range of domains. Most notably, this technique is combined with tree search to form an algorithm called Monte Carlo Tree Search (MCTS)~\cite{Browne2012Survey}, a method of finding the optimal decision in a given domain by taking a random sampling in the decision space and building a search tree accordingly \cite{Browne2012Survey}. Browne (et al.) further went into detail of explaining the Monte Carlo Tree Search, its variants and applications. 

Genetic programming has been used in conjunction with MCTS for classic strategy games such as \textit{Othello} and \textit{Dodgem}, as shown by Benbassat and Slipper \cite{benbassat2013evomcts}. They used each individual of the evolutionary algorithm as a function to evaluate a board position. This was then used during the roll-out portion of MCTS to select the action that would maximize the next board state in correspondence with the function \cite{benbassat2013evomcts}. Cazenave's work explored evolving the UCB1 equation for GO MCTS agents \cite{Cazenave_evolvingmonte-carlo}. Their work showed off a significant increase in performance, outperforming agents that utilized standard UCB and alternatives UCB1s designed specifically for GO. Similarly, Holmg\r{a}rd et al. successfully explored evolving the MCTS UCB function to create procedural personas for the game MiniDungeons 2 \cite{Holmgard2018automated}. For our purposes, we did not deploy the individuals in the fashion of Benbassat and Slipper. Rather, we used the functions as a means to select the \textit{promising node} to expand and the action of the best immediate child of the root node to return as shown in Holdgård's work \cite{Holmgard2018automated}.

\section{Match-3 Framework}
Match-3 games, a specific subset of the matching tile games family, are focused on for the experiments. The custom Match-3 framework, which is essentially a somewhat simplified version of Bejeweled, is built in Python 3, supports forward modeling, and uses a 7 by 7 game board. 
\subsection{Rules}
\begin{figure*}[t]
	\centering
    \begin{subfigure}{0.42\textwidth}
        \includegraphics[width=0.80\textwidth]{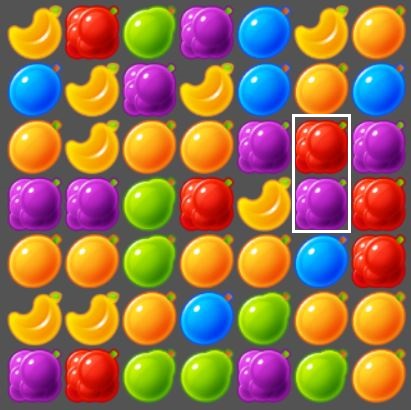}
        \subcaption{The white square shows a possible move a player can make to create a match of size 3.\newline}
        \label{fig:game}
    \end{subfigure}
    \begin{subfigure}{0.42\textwidth}
        \includegraphics[width=0.80\textwidth]{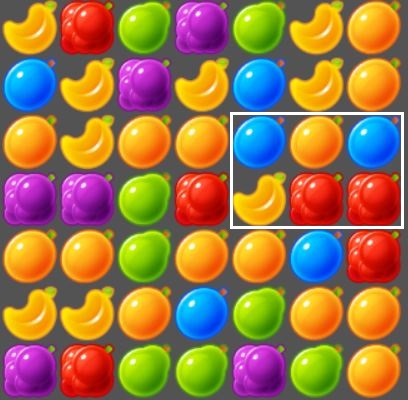}
        \subcaption{Pieces, as shown by the white square, fall in to replace empty spaces, and new pieces are introduced to keep the board filled.}
        \label{fig:game}
    \end{subfigure}
    \caption{Player makes a legal move to make a match of size 3}
    \label{fig:make_move}
\end{figure*}

\begin{figure*}[t]
	\centering
	\includegraphics[width=0.35\textwidth]{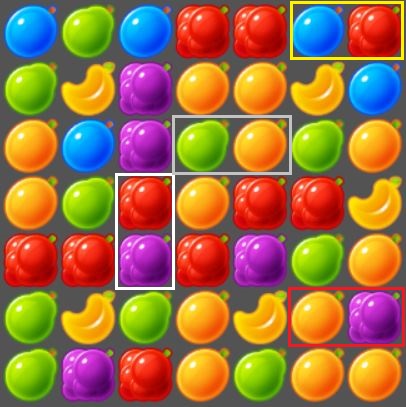}
    \caption{A few possible scenarios to make matches by swapping the 2 pieces in any of the highlighted colored squares.}
    \label{fig:many_moves}
\end{figure*}

The rules to play the custom Match-3 framework are similar to the rule-sets of games in the Match-3 paradigm. Given a board of N by M size, where N and M can be the same or different, swap two orthogonally adjacent cells to create a line of three or more identical cells. This is also referred to as a match of size 3 or more. If a swap does not make a match then the swap is undone, the board configuration is reset to the previous state, and no points are awarded. If a swap leads to a match, then the cells will be removed from the board, the cells above will fall down to fill the now empty spaces in the grid, as shown by Figure \ref{fig:make_move}. A match can be made either horizontally, vertically or both. The player is rewarded a corresponding number of points. If, after the board is refilled, another match of three or more identical cells exists then the process is repeated. However, this and all other proceeding matches from that single move are considered combos and points are rewarded based on a multiplier. The multiplier will be reset to 1 for each move the player makes. When a player makes a match of size greater than 3, they are as such rewarded a greater number of points. Further, for most Match-3 games, if four or five cells match, an additional power-up cell will be rewarded. For the purposes of the experiments, this last rule was disregarded. Figure \ref{fig:many_moves} shows different ways in which matches can be made. 

\subsection{Points and Score}
The fewest number of adjacent identical cells to trigger a match is three. As such, making this will reward the lowest number of points. 20 points are rewarded for each cell, and since there are 3 cells in the match, a total of 60 points is rewarded for the move that made the match of 3 identical cells. Using this as the base standard, the value of a cell increases by 10 for each additional cell in the match. The total number of points rewarded would be equal to the value of a cell multiplied by the number of value of a cell.
To account for the possibility of a combo being triggered an additional variable is introduced, a score multiplier. 
The score multiplier is initially equal to 1. Every time a combo is triggered the score multiplier is incremented by one. It is reset back to 1 once the board no longer has any matches and the user has to then make their next move.

\section{Methods}
Through MCTS, we were able to build an asymmetric unbalanced tree with a bias towards visiting nodes that performed to be more interesting based off the selection criteria and heuristic. Rather than using the standard Upper Confidence Bound 1 (UCB1) formula, we followed a strategy similar to that of the work of Christoff Holmg\r{a}rd (et. al.): to use genetic programming to evolve persona-specific evolution formulas \cite{Holmgard2018automated}. We replaced the standard node selection criteria in MCTS for genetically evolved player persona utility functions. 

\subsection{Procedural Personas}
Each procedural player persona agent had its own goal and as such the fitness of each was calculated differently. Agents 1 and 2 used the overall score after making a total of 20 turns as their fitness. Agent 1 looked to maximize the score, while Agent 2 looked to minimize it. Agent 1 used the returned score as the fitness of each individual in the population. Agent 2 took the score returned after playing 20 moves and negated the score for the fitness of each individual. This allowed for the strategy of obtaining an elitist, used later during the evolution of individuals of a generation, who focused on minimizing score. Agents 3 and 4 used the average length of legal available moves to make after a total of 20 turns. Agent 3 tried to maximize the overall average length while Agent 4 looked to minimize this value. Agent 3, similar to Agent 1, set the fitness of each individual equal to the average number of available moves returned after playing 20 moves. A similar strategy as used with Agent 2 was used with Agent 4 to minimize the total number of available moves. For all agents, since 50 simulations were played for each individual, the actual fitness is an average of the returned values of the 50 simulations (played by using the individuals associated equation as the replacement of the standard UCB function).

\subsection{Monte Carlo Tree Search}
As described above, MCTS is a tree search algorithm that, through biased selection of promising nodes, creates an unbalanced tree. For the purposes of our experiments we visited the root node 250 times, performed a rollout of initially 20 moves, and each MCTS agent performed a total of 20 real moves (aside from the simulations) on the actual game board. There was a negative linear correlation between the rollout length and the total number of moves the agent has made; as the number of actual moves the agent makes increased, the length of the rollout decreased. For example, if the agent made 4 real moves on the board, the rollout length when performing simulations would be $20-4$ or 16. To build the tree our agent performed the following procedure \cite{yannakakis2018artificial} \cite{Browne2012Survey}:

\subsubsection{\textbf{Selection}}
The most promising node to expand based upon the defined policy was selected, with a approach similar to that as explained in "Bandit Based Monte-Carlo Planning" by Kocsis (et al.)\cite{mcts_original}. For the vanilla MCTS agent, we used the Upper Confidence Bound 1 (UCB1) formula:
\begin{equation}
UCB1 = \overline{X_i} + C \ \sqrt[]{ \frac{2 * \ln {N^p_i}}{N_i} }
\end{equation}
where $\overline{X_i}$ is the average number of times node $i$ has won by achieving the defined goal, $N^p_i$ is the number of times the parent node has been visited, $N_i$ is the number of times the child node $i$ has been visited, $C$ is the exploration constant and set to $\frac{1}{\sqrt[]{2}}$.

\subsubsection{\textbf{Expansion}}
When a promising node was selected it represented a state in which other actions can still be taken to further progress in the game. A child array was created for the promising node holding all legal moves and corresponding states from taking these moves. A child was then taken at random to perform simulated rollout play.

\subsubsection{\textbf{Simulation}}
Once a child node was selected, simulated play was performed on the node. The actions taken were random and the number of actions taken, as described at the start of this subsection, initially started at 20 and decreased every time the search is performed.

\subsubsection{\textbf{Backpropogation}}
The results of the simulation step were backpropagated up the tree to each node from the selected child node for expansion to the root node.  

For our experiments, we focused on replacing the standard UCB1 equation during the selection of the most promising node and when selecting the best action to take. We instead used evolved mathematical formulas as explained in the previous section for the procedural persona agents.

\subsection{Evolutionary Policy} 
Genetic programming evolved discrete structures; mathematical formulas can be evolved by breaking down the representation of an equation into a syntax tree. This was denoted as the chromosome representation \cite{yannakakis2018artificial}. All the nodes of the tree contained either a \textit{binary} or \textit{unary} mathematical operation. The four binary functions were addition, multiplication, division, and subtraction, though only the former three were utilized in the actual equations for the experiments. The unary operator square root was also one of the mathematical operations used in the equations. All the leaf nodes were left to being either some predefined variable or a constant. Constant values were defined to be uniformly randomly generated floats within [0, 10]. Variables used include:
\begin{itemize}
\item number of times a child node of the current node has won the game
\item number of times a child node of the current node has been visited
\item number of times the current node has been visited
\item total number available moves of the current child node being evaluated
\end{itemize}

We utilized this approach to create an initial population of 100 unique individuals, meaning there were no duplicates when the equation for each individual was reduced and simplified. For each individual, the chromosome representation of the mathematical equation associated to the individual was of minimum depth 2 and maximum depth 6. 
When making the initial population, we first simplified the prospective individual's equation before checking it against all previous individuals currently accepted in the population. If there was a match then the prospective equation was disregarded, otherwise it was added to the population. Equivalence of two equations was defined as when two simplified equations subtracted from each other evaluated to 0. For example \textit{x-4} and \textit{-4+x} would be equivalent as \textit{(x-4) - (-4+x)} evaluates to 0. In this case, the equation that came later would be disregarded. When performing the MCTS, the UCB equation was completely replaced by each individual's equation as the evaluation heuristic for selecting the \textit{promising node} and the move of the best immediate child of the root node. 

After all 100 individuals finished playing 50 games each, results for the population were formulated and saved. The $\mu$ +$\lambda$ evolution strategy for genetic evolution was performed on the individuals in the following order: save the top 10\% of elitist, perform mutation and crossover on the remaining population \cite{yannakakis2018artificial}. For our purposes the elitist population results to the top 10 individuals based on their fitness and goal/criteria of player personas of which they were modeling, as explained previously. Of the remaining 90 spots, half was given for mutating individuals and the other half, the remaining 45 spots, was given to crossover. Mutation was performed and calculated first. 

\textbf{Mutation} is defined as taking a random chromosome and replacing it with another. A random sample of 45 individuals was chosen and there existed a 50\% chance of mutating a constant for each selected individual. Then the Deap Evolutionary Tools genetic programming mutUniform function\footnote{http://deap.readthedocs.io/en/master/api/tools.html\#deap.gp.mutUniform} was used on each of the 45 individuals. Before an individual was added to the population for the next generation, it was first compared against all existing individuals in the next generation population and disregarded if it was equivalent to any of those individuals. The same standard of equivalence, as previously defined, held. This process was repeated until a total of 45 possibly mutated individuals were added to the population for the next generation. After the mutation, crossover was performed to produce the remaining population. 

\textbf{Crossover} is when two random chromosomes from 2 selected individuals cross-over or swap to create two offspring. Using the current population, we created a randomly shuffled list of all possible pair combinations of individuals. Then while the number of next generations population size was under 100, performed crossover on selected pairs. If an offspring produced a duplicate of a preexisting individual using the defined equivalence test, the child would be disregarded.

We used a strategy in which every generation tried to out-perform the previous generation. It was that the highest fitness from the previous generation was saved and set as the goal for the next generation that was about to start playing games, the one that was just recently created following the given procedure. This strategy would progressively push each generation to try and out-perform the previous as they tried to reach a higher end goal within the Monte Carlo Tree Search, until a global maxima/minima was reached. At which point each generation would begin to score in roughly the same range.

\section{Experiments}
We ran a total of 4 experiments for the defined personas. For each experiment we followed the approach to randomize and save 50 seeds for each generation and use the same 50 seeds for each individual in the population. Each individual in the population would play a total of 50 games, 1 for each of the 50 seeds. For each game, the MCTS agent would make a total of 20 turns and perform its tree search for the action to take for each move. The fitness for each individual would be the average score of the scores from the 50 total games played by the individual. Then we re-used the seeds to have both Vanilla MCTS and Random agent play out games with these seeds and depending on the agent return different criteria. For Agents 1 and 2, the Vanilla and MCTS agent returned their final score which is then averaged for each generation of seeds. For Agents 3 and 4, the Vanilla and MCTS agent returned the average number of available moves also for 20 turns of game play, which is then similarly averaged for each generation of seeds .We then calculated the mean of all the generations for the Vanilla MCTS and Random Agents and plotted those 2 values, one for the Vanilla MCTS Agent and the other for the Random Agent, for Figures 3 through 6.

\begin{figure}[t]
	\centering
    \includegraphics[width=\linewidth]{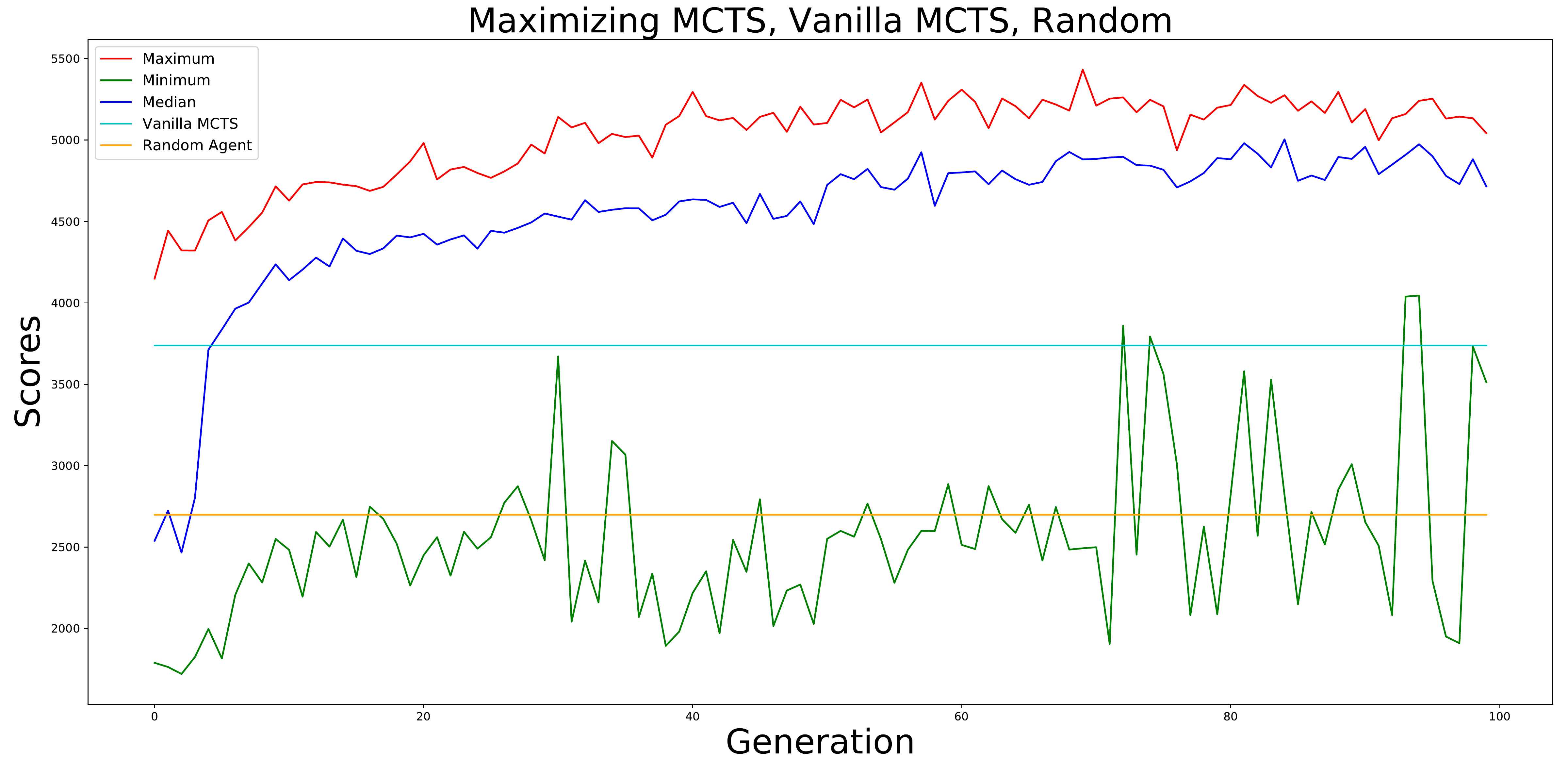}
    \caption{Experiment 1. Maximizing score over 100 generations with a population size of 100, 50 games per individual and 20 moves per individual.}
    \label{fig:max_score}
\end{figure}

\begin{figure}[t]
	\centering
    \includegraphics[width=\linewidth]{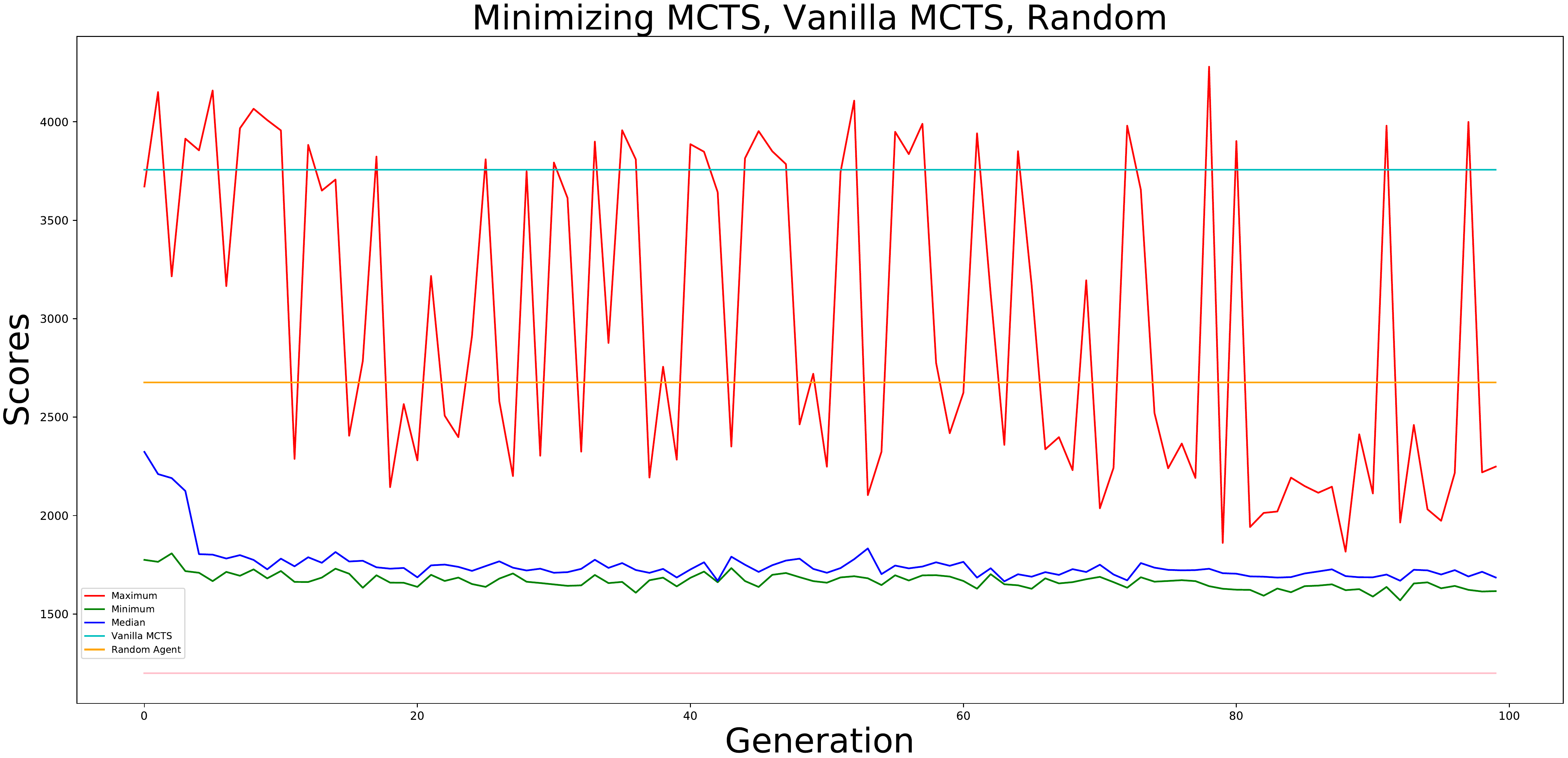}
    \caption{Experiment 2. Minimizing score over 100 generations with a population size of 100, 50 games per individual, and 20 moves per individual.}
    \label{fig:min_score}
\end{figure}

\begin{figure}[t]
	\centering
    \includegraphics[width=\linewidth]{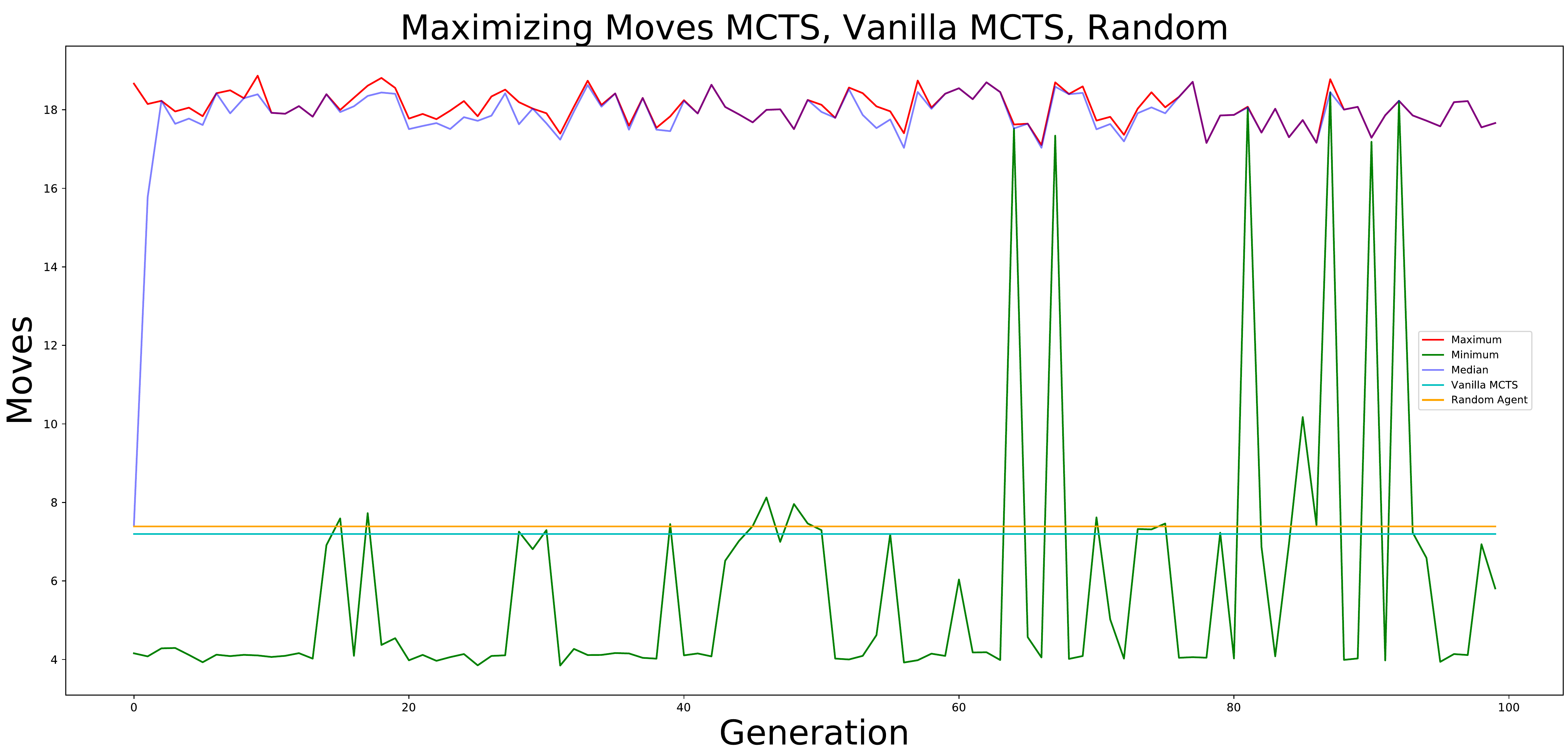}
    \caption{Experiment 3. Maximizing average number of moves over 100 generations with a population size of 100, 50 games per individual and 20 moves per individual.}
    \label{fig:max_move}
\end{figure}

\begin{figure}[t]
	\centering
    \includegraphics[width=\linewidth]{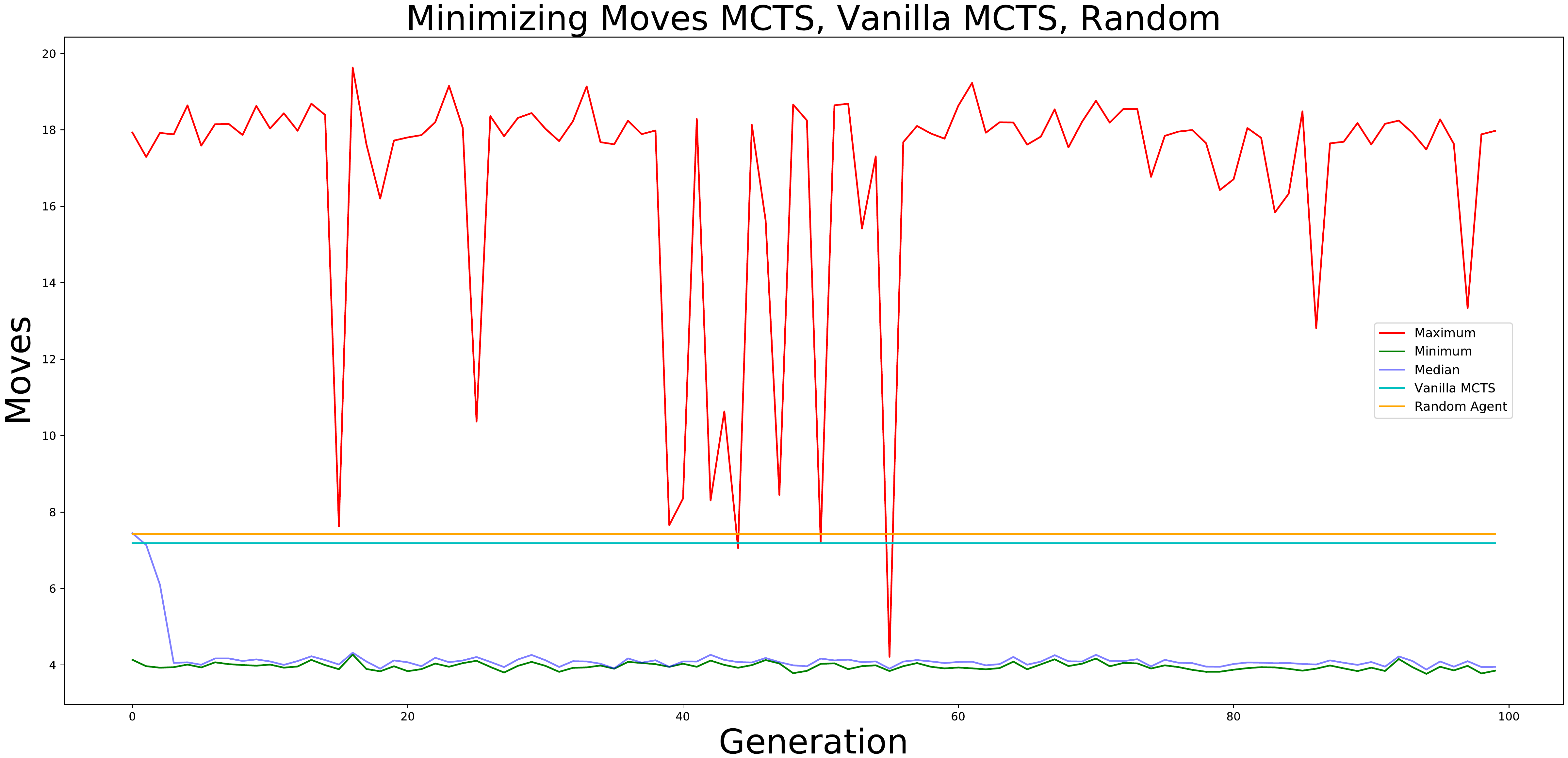}
    \caption{Experiment 4. Minimizing average number of moves over 100 generations with a population size of 100, 50 games per individual, and 20 moves per individual. The pink line represents the theoretically achievable lowest possible minimum score after 20 moves.}
    \label{fig:min_move}
\end{figure}

Figure \ref{fig:max_score} shows the results of the maximizing procedural persona agent. Figure \ref{fig:min_score} shows the results of the minimizing procedural persona. It is important to note that the lowest possible score to achieve in the framework is 1200 points, denoted as the straight pink line in the figure. 
Figure \ref{fig:max_move} shows the results for an agent maximizing the average number of available moves over 20 turns, while Figure \ref{fig:min_move} shows the results for minimizing the average number of available moves. 

In the following sections, we describe the results of the experiments, comparing them to the standard UCB1 and Random playing agents for each experiment, and to the results of the user study.

\section{Discussion}
For all experiments, noise produced between generations can be a result of fluctuations from individuals in the generation with the weakest fitness. Since we disregarded low performing individuals during the evolution, there lies a high probability  a few of the new individuals introduced into the next generation's population perform drastically different from the current generation's lowest performing individuals. 

Figure \ref{fig:max_score} reflects an overall increase in the agent's performance as the median quickly approaches the maximum for each generation, to the point where it begins to level out with some residual noise. The leveling off indicates that a maxima, possibly local maxima, is reached for the performance of trying to maximize the total score after making 20 moves per game.

Figure \ref{fig:min_score} reveals there are ways to play the game in which you perform worse than if you were to just play randomly. The minimum and median for each generation begin to level off roughly in the 1600 and 1700 point range, the possible playable global minima. The 400 to 500 point gap between the global minima and lowest possible number of points a player can make, which is 1200, indicates that there are unavoidable situations where a combo is forced to happen causing one to gain more points than the bare minimum points for a single turn.

Figure \ref{fig:min_move} shows that on average a player will have more than 1 move readily available. Figure \ref{fig:max_move} reveals the possibility for players to conduct a strategy in which they try to maximize the available number of moves in hopes of setting up the board. This grants the user more freedom and choice when deciding moves, and to focus on triggering combos by making one match and having pieces fall into matches that follow.

When comparing the evolution of the score maximizing agent from figure~\ref{fig:max_score} with that of the moves maximizing agent from figure~\ref{fig:max_move} we can spot considerable differences, the main one being the rate at which the different populations converge. While maximizing score slowly and constantly improves, maximizing moves peaks very early. This can be explained due to the fact that maximizing moves is an objectively easier strategy to execute than maximizing score. Increasing the number of moves available resorts to selecting the move that will create the highest number of possible moves on the next turn. While the optimization of such a strategy still requires multiple steps (e.g. thinking multiple steps ahead can help you set up a bigger payoff over multiple turns) in favor of a one-step look ahead, it does not rely on combos, opting to avoid making moves that will create them instead. These moves are arguably the hardest ones to optimize for, as well as the ones that impact your score the most.

\section{User study}
An online user study was conducted in which a total of 41 participants completed 6 rounds of the match-3 game. Each round consisted of 20 moves. Of the 6 rounds, 3 rounds used predetermined boards and falling pieces while the remaining 3 were completely randomized. The order of the 6 boards was randomized for each user.


Before starting the study, participants were asked questions regarding their profile. In our study, we had 41 participants: 29 males, 9 females, and 3 non-disclosed gender. 89.6$\%$ of males and 66.6$\%$ of females fell in the age range of 18-24. ~37.9$\%$ of males play games everyday and ~37.9$\%$ of males play games several times a week, 24$\%$ of males had never played a match-3 game, ~34.5$\%$ have played less than 10 matches of a match-3 game. ~44$\%$ of females play games once a month, ~44$\%$ of females play games several times a week, ~22$\%$ of females have played less than 10 matches, ~33$\%$ of females have played between 10-19 matches, and ~33$\%$ of females have played over 100 matches of a match-3 game.
\begin{table}[ht]
\label{tab:user_study_res}
\begin{center}
\caption{Result score statistics for users from user study for the 3 preset boards and 3 randomly generated boards.}
\resizebox{\linewidth}{!}{%
\begin{tabular}{|c|c|c|c|c|}
\hline
& Board 1 & Board 2 & Board 3 & Avg of 3 Random Boards \\
\hline
Average & 4530.24 & 3042.93 & 2911.71 & 3275.64 \\
\hline
Maximum & 7680 & 5260 & 6440 & 7060 \\
\hline
Minimum & 2120 & 1740 & 1740 & 1700 \\
\hline
\end{tabular}%
}
\end{center}
\end{table}

\begin{table}[ht]
\label{tab:user_study_res}
\begin{center}
\caption{Agent results for preset boards. Agent MaxS and MinS are the maximizing and minimizing score agents, while Agent MaxM and MinM are the maximizing and minimizing average number of available move agents, respectively. The score from the top performer of the final generation playing the boards is shown. In addition, the average number of moves for agents MaxM and MinM are also shown in parenthesis. Vanilla represents the Vanilla MCTS and Random represents an agent choosing moves at Random. }
\resizebox{\linewidth}{!}{%
\begin{tabular}{|c|c|c|c|c|c|c|}
\hline
& Agent MaxS & Agent MinS & Agent MaxM & Agent MinM & Vanilla & Random \\
\hline
Board 1 & 7080 & 1740 & 2640 (14.35) & 2720 (4.65) & 5240 & 2720 \\
\hline
Board 2 & 6460 & 1380 & 1560 (12.5) & 2120 (3.75) & 4500 & 3840 \\
\hline
Board 3 & 7040 & 1500 & 2160 (12.35) & 1680 (3.25) & 5840 & 2340 \\
\hline
\end{tabular}%
}
\end{center}
\end{table}

Results for the user study are shown in Table 1. Results for the agents playing the same three preset boards are shown in Table 2. Players on average outscore every persona agent but MaxS. The MinS agent manages to have a lower score than even the lowest scoring user in all boards. Meanwhile, MaxS agent outscores the highest scoring user in all but one board, in which it comes very close. This leads us to observe that using MaxS and MinS can provide a good score interval for a stage, emulating high and low performance respectively. Overall, Vanilla MCTS has a better performance than average users, meaning it is still a powerful algorithm to playtest the game with.

Another point to notice is the average scores between different boards. Board 1 has higher scores in all scenarios, for both players and persona agents, when compared to Boards 2 and 3. This indicates that it is an easier stage to play, which is valuable information when trying to balance the game.


\section{Conclusion and future work}
In this paper, we presented a procedure to formulate and genetically evolve mathematical equations to represent various play-styles for Match-3 games. We developed an agent that mimics a long term human player who looks to strategically optimize the maximum number of points that can be achieved through a series of actions after a certain number of moves. We additionally evolved an agent that aimed to minimize its overall score. From this, it shows the possibility of being able to perform worse than simply playing the Match-3 game randomly and that receiving a combo is nearly unavoidable. By deploying these agents into real world Match-3 games, it opens up the ability to analyze level designs and the approaches taken to play levels by various player perspectives. 

By using such agents we were able to extract features from pre-made stages. Our score maximizing and score minimizing agents allowed us to evaluate and estimate the range of performance for human players. Also, comparing the performance of such agents across multiple boards aided in measuring what can be perceived as their difficulty levels. These findings were supported by the player data we collected in our user study.

For future work, we propose developing a reinforcement learning algorithm that could use the collected user data to simulate human behavior in decision making. Another avenue to explore would be to modify the Match-3 engine to include special pieces that form from different combinations of matches greater than 3. We believe that introducing these special candies will allow for a greater variation in skill based on how they are utilized in the level. It would be interesting to observe any changes in the performance of the four agents, as they may perform and make decisions differently after the introduction of these special pieces.


\bibliographystyle{IEEEtran}
\bibliography{bibliography}

\begin{thebibliography}{10}
\providecommand{\url}[1]{#1}
\csname url@samestyle\endcsname
\providecommand{\newblock}{\relax}
\providecommand{\bibinfo}[2]{#2}
\providecommand{\BIBentrySTDinterwordspacing}{\spaceskip=0pt\relax}
\providecommand{\BIBentryALTinterwordstretchfactor}{4}
\providecommand{\BIBentryALTinterwordspacing}{\spaceskip=\fontdimen2\font plus
\BIBentryALTinterwordstretchfactor\fontdimen3\font minus
  \fontdimen4\font\relax}
\providecommand{\BIBforeignlanguage}[2]{{%
\expandafter\ifx\csname l@#1\endcsname\relax
\typeout{** WARNING: IEEEtran.bst: No hyphenation pattern has been}%
\typeout{** loaded for the language `#1'. Using the pattern for}%
\typeout{** the default language instead.}%
\else
\language=\csname l@#1\endcsname
\fi
#2}}
\providecommand{\BIBdecl}{\relax}
\BIBdecl

\bibitem{Holmgard2018automated}
C.~Holmg\r{a}rd, M.~C. Green, A.~Liapis, and J.~Togelius, ``Automated
  playtesting with procedural personas through {MCTS} with evolved
  heuristics,'' \emph{CoRR}, vol. abs/1802.06881, 2018.

\bibitem{Holmgard2014evolving}
C.~Holmg\r{a}rd, A.~Liapis, J.~Togelius, and G.~N. Yannakakis, ``Evolving
  personas for player decision modeling,'' in \emph{2014 IEEE Conference on
  Computational Intelligence and Games (CIG)}, Aug 2014, pp. 1--8.

\bibitem{tastan2011learning}
\BIBentryALTinterwordspacing
B.~Tastan and G.~Sukthankar, ``Learning policies for first person shooter games
  using inverse reinforcement learning,'' in \emph{Proceedings of the Seventh
  AAAI Conference on Artificial Intelligence and Interactive Digital
  Entertainment}, ser. AIIDE'11.\hskip 1em plus 0.5em minus 0.4em\relax AAAI
  Press, 2011, pp. 85--90. [Online]. Available:
  \url{http://dl.acm.org/citation.cfm?id=3014589.3014604}
\BIBentrySTDinterwordspacing

\bibitem{Alessandro2008Defining}
\BIBentryALTinterwordspacing
A.~Tychsen and A.~Canossa, ``Defining personas in games using metrics,'' in
  \emph{Proceedings of the 2008 Conference on Future Play: Research, Play,
  Share}, ser. Future Play '08.\hskip 1em plus 0.5em minus 0.4em\relax New
  York, NY, USA: ACM, 2008, pp. 73--80. [Online]. Available:
  \url{http://doi.acm.org/10.1145/1496984.1496997}
\BIBentrySTDinterwordspacing

\bibitem{Alessandro2009patterns}
A.~Drachen and A.~Canossa, ``\BIBforeignlanguage{English}{Patterns of play:
  Play-personas in user-centred game development},'' in
  \emph{\BIBforeignlanguage{English}{Proceedings of DiGRA 2009}}.\hskip 1em
  plus 0.5em minus 0.4em\relax DIGRA, 2009.

\bibitem{holmgard2014generative}
C.~Holmg\r{a}rd, A.~Liapis, J.~Togelius, and G.~N. Yannakakis, ``Generative
  agents for player decision modeling in games,'' in \emph{Poster Proceedings
  of the 9th Conference on the Foundations of Digital Games (FDG)}, 2014.

\bibitem{Browne2012Survey}
C.~Browne, E.~Powley, D.~Whitehouse, S.~Lucas, P.~Cowling, P.~Rohlfshagen,
  S.~Tavener, D.~Perez~Liebana, S.~Samothrakis, and S.~Colton, ``A survey of
  monte carlo tree search methods,'' \emph{IEEE Transactions on Computational
  Intelligence and AI in Games (TCIAIG)}, vol. 4:1, pp. 1--43, 03 2012.

\bibitem{benbassat2013evomcts}
A.~Benbassat and M.~Sipper, ``Evomcts: Enhancing mcts-based players through
  genetic programming,'' in \emph{2013 IEEE Conference on Computational
  Intelligence in Games (CIG)}, Aug 2013, pp. 1--8.

\bibitem{Cazenave_evolvingmonte-carlo}
T.~Cazenave, ``Evolving monte-carlo tree search algorithms,” dept,''
  \emph{Inf., Univ. Paris}, p. 2007.

\bibitem{yannakakis2018artificial}
G.~N. Yannakakis and J.~Togelius, \emph{{Artificial Intelligence and
  Games}}.\hskip 1em plus 0.5em minus 0.4em\relax Springer, 2018,
  \url{http://gameaibook.org}.

\bibitem{mcts_original}
L.~Kocsis and C.~Szepesv{\'a}ri, ``Bandit based monte-carlo planning,'' in
  \emph{Machine Learning: ECML 2006}, J.~F{\"u}rnkranz, T.~Scheffer, and
  M.~Spiliopoulou, Eds.\hskip 1em plus 0.5em minus 0.4em\relax Berlin,
  Heidelberg: Springer Berlin Heidelberg, 2006, pp. 282--293.

\end{thebibliography}

\end{document}